\documentclass[sigconf]{acmart}
\usepackage{caption}
\usepackage{subcaption}

\renewcommand\footnotetextcopyrightpermission[1]{} 
\def\BibTeX{{\rm B\kern-.05em{\sc i\kern-.025em b}\kern-.08emT\kern-.1667em\lower.7ex\hbox{E}\kern-.125emX}}
    
\acmConference[TR]{TR}{2020}{Alibaba DAMO Academy}

\begin{document}

%
\title[{RobustTAD}]{RobustTAD: Robust Time Series Anomaly Detection via Decomposition and Convolutional Neural Networks }

%

\author[Gao et al.]{Jingkun Gao, Xiaomin Song, Qingsong Wen, Pichao Wang, Liang Sun, Huan Xu}
\email{{jingkun.g, xiaomin.song, qingsong.wen, pichao.wang, liang.sun, huan.xu}@alibaba-inc.com}
\affiliation{%
  \institution{Machine Intelligence Technology, DAMO Academy, Alibaba Group Inc.}
}

\begin{abstract}
The monitoring and management of numerous and diverse time series data at Alibaba Group calls for an effective and scalable time series anomaly detection service. In this paper, we propose RobustTAD, a Robust Time series Anomaly Detection framework by integrating robust time series decomposition and convolutional neural network for time series data. The time series decomposition can effectively handle complicated patterns in time series, and meanwhile significantly simplifies the architecture of the neural network, which is an encoder-decoder architecture with skip connections. This architecture can effectively capture the multi-scale information from time series, which is very useful in anomaly detection. Due to the limited labeled data in time series anomaly detection, we systematically investigate data augmentation methods in both time and frequency domains for the decomposed components. We also introduce label-based weight and value-based weight in the loss function by utilizing the unbalanced nature of the time series anomaly detection problem. Compared with the widely used forecasting-based anomaly detection algorithms, decomposition-based algorithms, traditional statistical algorithms, as well as recent neural network based algorithms, our proposed RobustTAD algorithm performs significantly better on public benchmark datasets. It is deployed as a public online service and widely adopted in different business scenarios at Alibaba Group. 



\end{abstract}




%

%

\maketitle

\section{Introduction}

As the rapid increase of time series data due to the developments of Internet of Things (IoT) and many other connected data sources, real-time time series anomaly detection is a required capability in many real-world applications, such as predictive maintenance, intrusion detection, fraud prevention, cloud platform monitoring and management, business data monitoring etc. For example, in the monitoring of a data center, usually we need to monitor metrics from different levels, ranging from physical machine, docker, to service on each docker. Another example is the tracking of sales amount as some unusual high amounts may be caused by some cheating transactions. In this paper we focus on the anomalies in a univariate time series. Specifically, we define an anomaly in a time series as an observation which is significantly different from previous normal observations, which we call its ``context". 
Formally, given a sequence $\cdots, x_{t-w}, x_{t-w+1}, \cdots, x_{t-1}, x_t$, our goal is to determine whether the newest observation $x_t$ is an anomaly by considering the current and previous observations up to a window size $w$. 


Time series anomaly detection has been researched for over a long time~\cite{AD_survey_2009,AD_gupta2014outlier}. However, no existing
algorithm can work well in large-scale applications. Here we summarize some challenges of anomaly detection for time series. Firstly, the time series data in real-world scenarios is quite diverse. It does not only contain the temporal dependency, but may also exhibit more complicated patterns, such as abrupt change of trend and seasonality shift and fluctuation. Secondly, the labeling of anomalies is challenging and thus very limited labeled data is available. In time series anomaly labeling, we need to compare each sample with its previous context, thus it is quite time-consuming to label all samples in a time series manually. In addition, the definitions of anomaly may vary in different scenarios. Thirdly, almost all time series anomaly detection system is required to respond in real time. Therefore, the anomaly detection algorithm should be efficient and can handle a large number of metrics in parallel with low latency. 


Recent years witness the advances in deep learning~\cite{deep:learning:book}. Compared with the wide success of deep learning in computer vision and natural language processing, it only has limited applications in time series. When processing time series data, a key challenge is how to consider temporal dependency and extract features from the original time series effectively. In particular, in time series anomaly detection the multi-scale information is very important as it effectively helps to define the context so anomaly can be compared and determined. When more complex patterns in seasonality and trend are present in time series, it is even more challenging to build a deep learning model. In addition, deep learning models generally require a lot of labeled data for training. Thus, data augmentation is a crucial step in model training. Unlike data augmentation for images and speeches\cite{Um:data:aug}, few work has been done on data augmentation for time series. Specifically, as recurrent neural network (RNN) and long short-term memory (LSTM) network are ideal tools to model temporal dependency, several works~\cite{Malhotra:LSTM-Ad:2015,Kim:LSTM-Ad:2016} have been proposed for time series anomaly detection. However, a challenge for these networks is how to deal with the seasonality especially long seasonality in a general manner. For example, \cite{Lai:LSTM:forecasting:2018} proposes to create a shortcut connection to learn seasonality directly, but its generality is harmed by time series data with different seasonality lengths. In contrast, convolutional neural networks (CNN) only has limited applications in time series classification and clustering~\cite{Yang:CNN-ts-classification:2015}, and recently in anomaly detection~\cite{Wen2019}. How to encode the temporal dependency and complicated time series patterns in CNN still remains an open problem.




In this paper, we propose RobustTAD, a Robust Time series Anomaly Detection framework integrating time series decomposition with convolutional neural network. 
In the time series decomposition, we first apply our RobustPeriod~\cite{WenRobustPeriod20} algorithm to detect if the time series is periodic and estimate its period length. Based on the periodicity, we apply either our RobustSTL~\cite{Wen2019} (an effective seasonal-trend decomposition algorithm for periodic time series) or our RobustTrend~\cite{WenRobustTrend19} (an effective trend filtering algorithm for non-periodic time series), to decompose the input time series into different components. 
Then we build a convolutional neural network based on the decomposed components to predict anomaly. Our network is based on U-Net~\cite{Ronneberger2015}, an encoder-decoder architecture with skip connection to extract multi-scale features from time series. Performing time series decomposition for time series before training the network has several benefits: 1) as the seasonality component is explicitly extracted, there is no need to build complex structure in the neural networks to deal with it, which significantly simplifies the network structure and meanwhile improves model performance; 2) it leads to a more general anomaly detection framework for time series with different characteristics, such as different seasonal lengths, different types of trend, etc. As limited labeled data is available in typical anomaly detection tasks, data augmentation is crucial to the successful training of the neural network. In this paper, we systematically investigate the data augmentation methods for time series, in both time domain and frequency domain for decomposed components. By utilizing the unbalanced nature of time series anomaly detection problem, we adjusted the loss function by introducing the label-based weight and value-based weight. To the best of our knowledge, our work is the first one which integrates the time series decomposition with deep neural networks for time series anomaly detection. Our experiments on  public benchmark datasets show that our integrated framework RobustTAD outperforms other popular algorithms, including forecasting-based methods, decomposition-based methods, and methods based deep learning but without time series decomposition.

\vspace{-0.4cm}
\section{Related Work}

The pioneer study of time series anomaly detection started in 1972, where Fox first adopted an auto-regressive forecasting model and devised specific statistical tests on forecasting errors to detect anomalies~\cite{Fox1972}. Since then, the time series anomaly detection problem has drawn attentions in statistics~\cite{Burman1988,Vallis2014,Siffer2017}, data mining~\cite{Keogh2002,Muthukrishnan2004,Sun2006} and machine learning communities~\cite{Laptev2015,Lavin2016,Guha2016,Xu2017}. A detailed review of time series anomaly detection can be found in~\cite{AD_survey_2009,AD_gupta2014outlier}. Here we summarize several of proposed approaches with two critical steps: one is the \textbf{representation learning}, and the other is the \textbf{detection}. 

The purpose of the \textbf{representation learning} is to learn a new representation so the anomalous points stand out, as the raw time series usually contain complex seasonality, trend, and noisy patterns, making it difficult to be used as the context to identify anomalies. As a result, it is a typical choice to transform the raw time series into another feature space where each point can be identified more obviously based on simplified context. For example, it is much easier to identify a small spike anomaly from a de-seasoned and de-noised signal compared with detecting this anomaly point from the origin time series having complex seasonality and noises.

The most popular representation is the forecasting error from a forecasting model for the time series. These forecasting models include ARIMA~\cite{Fox1972,Burman1988}, Bayesian method~\cite{Xu2017}, sequence memory algorithm hierarchical temporal memory~\cite{Ahmad2017}, and more recently LSTM based neural networks~\cite{Hundman2018,Zhang2019}. Due to the off-the-shelf availability of many forecasting packages, these forecasting algorithms attract a lot of attentions in both industry and academy. However, many of them are not robust to noises or outliers, as the forecasting model is built upon both normal and abnormal data, leading to unsatisfying detection performance when the training data contains portion of anomaly points. 




Another alternative approach involves applying decomposition on raw time series and utilizes the decomposed residuals or trends as the new representation for detection~\cite{Vallis2014,Hochenbaum2005,Choudhary2018,wang2019online}. There are also researchers who combine both representations, utilizing the decomposition results to facilitate forecasting tasks~\cite{Choudhary2018}. 
However, one of the major challenges in decomposition-based anomaly detection is how to detect seasonality and remove the seasonality/trend component that is commonly observed in many real-world data. 
In our previous work, for periodicity detection, we have already shown that our proposed RobustPeriod~\cite{WenRobustPeriod20} is better than the existing methods such as {SAZED}~\cite{Toller2019} and {AUTOPERIOD}~\cite{cPD_vlachos2005Autoperiod}; for seasonal-trend decomposition, our proposed RobustSTL~\cite{RobustSTL_wen2018robuststl} is better than the existing methods such as STL~\cite{STL_cleveland1990stl}, TBATS~\cite{TBATS_de2011forecasting}, and STR~\cite{STR_dokumentov2015str}; while for trend filtering, our proposed RobustTrend~\cite{WenRobustTrend19} is better than the existing methods such as $\ell_1$ trend filter~\cite{cmp_l1hpemp} and {mixed trend filter}~\cite{tibshirani2014adaptive}.
It is observed that conducting time series analysis on decomposed signals has shown improved performance on many tasks, such as forecasting, anomaly detection, etc. Besides the most commonly known forecasting and decomposition approaches for representation learning, there are other machine learning based approaches, for example, ~\cite{Guha2016} uses the complexity of the tree-based model to encode the anomaly, ~\cite{xu2018unsupervised,su2019robust} examine the reconstruction probability of the variational autoencoder to identify anomalies, and~\cite{Ren2019} utilizes salience detection techniques from computer vision domain to transform the original time series.

In the step of \textbf{detection}, we can adopt both unsupervised and supervised methods. In unsupervised setting, a straightforward but effective method is setting an empirically chosen threshold~\cite{Laptev2015,Guha2016,Hundman2018,xu2018unsupervised,Zhang2019,Ren2019,su2019robust} on the new representation. Besides the simple thresholding method, many statistical tests based approaches have also been used on the learned representation~\cite{Fox1972,Burman1988,Vallis2014,Ahmad2017,Xu2017,Hochenbaum2005,Choudhary2018,Adhikari2019}. The most widely known ``3-sigma'' rule is essentially a two-sided one sample t-test with a pre-chosen significance level. Nevertheless, challenges arise when a suited threshold need to be chosen, or a sensitivity parameter like a significance level need to be picked. Given a few labels from users' feedback, it is possible to tune these parameters accordingly. However, as more and more data become available incorporating diverse patterns and the challenges of learning a good representation, the simple threshold methods or statistical tests will fall short to model such complexity in time series. Recently, some supervised methods have been proposed~\cite{Ren2019,Ye2019,Wen2019}. These approaches feed either raw time series or the new representation into the network, and utilize different architectures to train a neural network to identify anomalies in time series. Some recent attempts~\cite{Wen2019} which apply network architectures from other domains directly only produce sub-optimal performance.

\section{Methodologies}\label{subsec:framework}

In this section, we introduce the proposed time series anomaly detection framework taking advantage of both decomposition and deep convolutional neural network. 

\subsection{Decomposition}
Usually the time series data contains different components. In this paper, we assume a time series can be decomposed as the sum of trend, seasonality, and remainder components: 
\begin{equation*}
  x_t = \tau_t + s_t + r_t, t=1,2,\cdots,N  
\end{equation*}
where $x_t$ denotes the original signal at time $t$, $\tau_t$ denotes the trend, $s_t$ denotes the seasonality if the time series is periodic and $r_t$ is the remainder component. 


Note that many decomposition algorithms have been proposed in the literature to perform time series decomposition, such as STL~\cite{STL_cleveland1990stl}, TBATS~\cite{TBATS_de2011forecasting}, and STR~\cite{STR_dokumentov2015str} for periodic time series, and $\ell_1$ trend filter~\cite{cmp_l1hpemp}, {mixed trend filter}~\cite{tibshirani2014adaptive} for non-periodic time series.
Although they are popular and effective in many tasks especially in forecasting, they do not perform well in anomaly detection due to the following reasons: 1) most of them cannot capture the abrupt change of trend and remainder in time, but these abrupt changes often correspond to anomalies and should be detected; 2) seasonality shift and fluctuation occur frequently in many real-world scenarios; 3) not robust to outliers and noises; 4) inefficiency to deal with time series with long seasonal length. 


In this paper, we apply our previous work RobustPeriod~\cite{WenRobustPeriod20} algorithm to detect if the time series is periodic and estimate its period length. For periodic time series, we adopt our previous work RobustSTL~\cite{RobustSTL_wen2018robuststl} to properly deal with all aforementioned challenges. It consists of 4 main steps, including noise removal, trend extraction, seasonality extraction and final adjustment. Note that it is an alternating algorithm as it is challenging to separate the trend component and the seasonality component directly. Specifically, the bilateral filtering~\cite{bilateral:book} is introduced to denoise the input time series. In the trend extraction step, we assume the seasonality component is already extracted (in the beginning we assume the seasonality component is strictly repeated in different seasonalities), and formulate the trend estimation as a regression problem. In this regression problem, the least absolute deviation (LAD) loss function is introduced to handle outliers robustly. In particular, to handle both the slow and abrupt change of trend, the $\ell_1$-norm regularization of the first and second difference of the trend is incorporated. After the trend component is updated, we estimate the seasonality component by extending bilateral filtering to the seasonal manner so seasonality shift and fluctuation can be handled properly. Finally the three components, i.e., trend, seasonality, and remainder, are adjusted to ensure that the sum of seasonality component in a seasonal period equals to zero. These components could be used as inputs to the designed network. Similarly, for non-periodic time series, we adopt our previous work RobustTrend~\cite{WenRobustTrend19} filter to obtain the trend and remainder components robustly and efficiently.
In this paper, we focus on the remainder component as the input to the neural network.

\subsection{Encoder-Decoder Network}


\subsubsection{Architecture}

Time series anomaly detection is a point-wise dense prediction problem. In other words, for a time series $\mathbf{x} = \{x_t\}_{t=1}^N$, the goal is to produce a sequence of the same length $\mathbf{y} = \{y_t\}_{t=1}^N$ where $y_i\in\{0,1\}$ denotes whether $x_i$ is an anomaly or not. Notice this dense problem shares many similarities with the image segmentation problem in computer vision where a pixel-wise inference is needed. As is reviewed in~\cite{taghanaki2019deep}, many state-of-the-art image segmentation approaches have adopted an encoder-decoder network structure, which helps to extract hidden features from the input. Meanwhile, to encode the complex patterns of the time series, it is necessary to consider both the local and global information (or multi-scale feature), which leads to the use of a skip connection that preserves the local information from the encoder layer by concatenating to the decoder input.

\begin{figure}[!htb]
    \centering
    \includegraphics[width=1\linewidth]{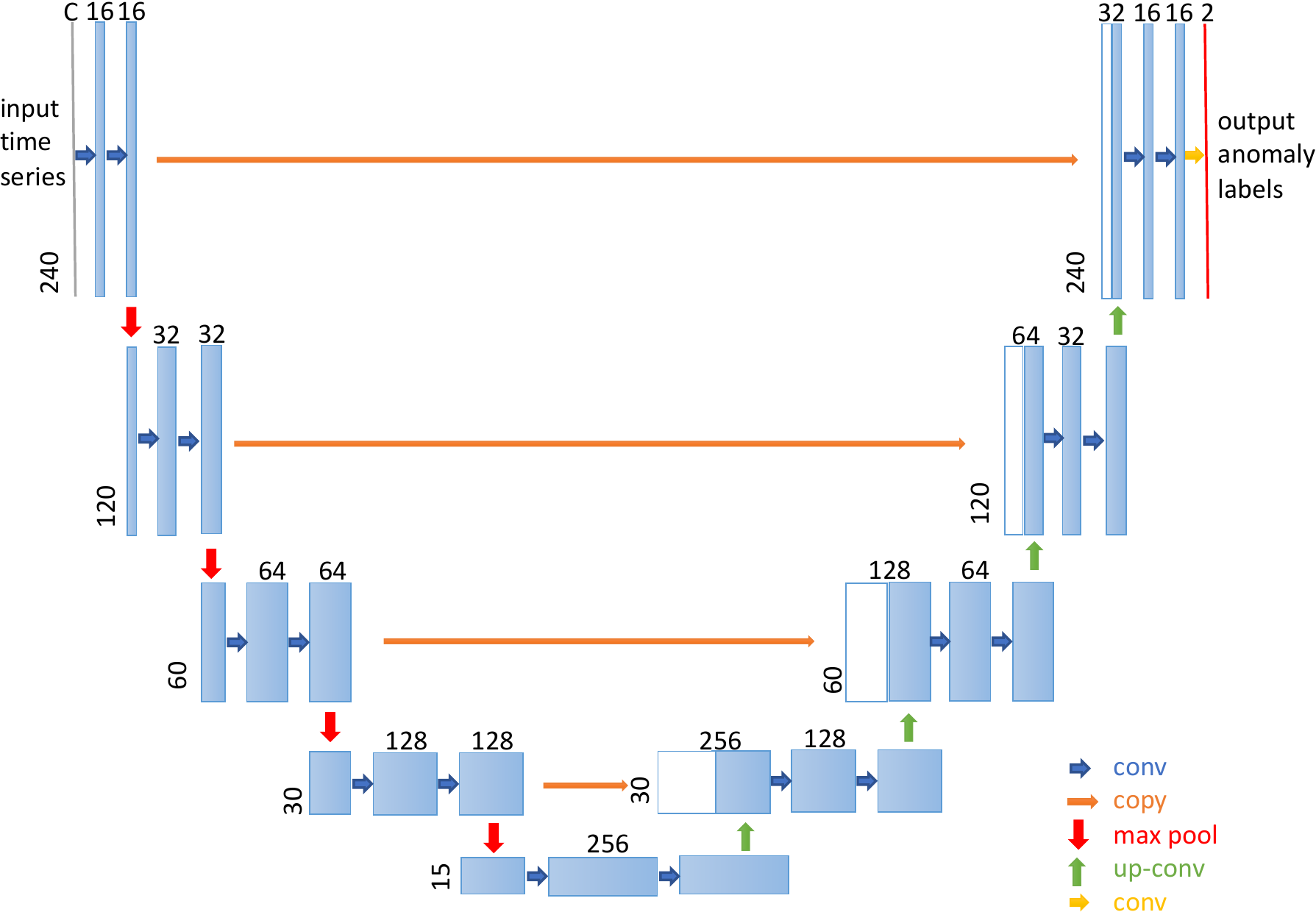}
    \caption{Architecture of Encoder-Decoder network for time series anomaly detection}
    \label{fig:3-net-arch}
\end{figure}

As a result, we adopt an encoder-decoder network architecture with skip connections, as known as U-Net structure~\cite{Ronneberger2015} shown in Figure~\ref{fig:3-net-arch}. The network is trained by feeding multiple samples of time series segment along with the corresponding labels for that segment. It is worth mentioning that utilizing U-Net structure for time series anomaly detection has shown promising results in~\cite{Wen2019}. However, in this paper, we would like to show that applying a U-Net structure directly without modification would still lead to a sub-optimal performance. Many adjustments are needed to achieve the optimal results, as is described in the following subsections.

\subsubsection{Weight Adjusted Loss}
As we cast the time series anomaly detection problem as a supervised classification problem, one must be aware that the problem is highly unbalanced. In other words, for a long time series with thousands of data points, there might only be a few anomalies. Applying the most commonly used pixel-wise cross entropy loss from U-Net would lead to unsatisfying performance as equal weights are assigned to normal samples and anomalous samples, yet the few anomalous samples that contribute most to training the model have been overshadowed by the vast majority of normal samples. It is natural to realize that we should assign more weights to anomalous samples. We refer this as \textbf{label-based weight}. Additionally, we also notice the value of the points itself should also be treated differently. In other words, for a point which is different from its neighbours, it is supposed to be more important to the model compared with the points which are all same, since they are more likely to be anomalous. This can be achieved by generating weights based on difference between its neighbours. We refer this \textbf{value-based weight}.
More specifically, for a time series $\mathbf{x} = \{x_t\}$, with ground truth labels $\mathbf{y} = \{y_t\}$, we denote the predictions from the network as $\hat{\mathbf{y}} = \{\hat{y}_t\}$, which are essentially the probabilities of $x_t$ being anomalies after the soft-max layer. The original cross-entropy loss is written as

$$
Loss(\hat{\mathbf{y}}, \mathbf{y})= -\sum_{t}y_t\log(\hat{y}_t) + (1-y_t)\log(1-\hat{y}_t)
$$

By applying the adjustments on two different scales, we can write the new weight adjusted loss as

$$
WALoss(\hat{\mathbf{y}}, \mathbf{y})= -\sum_{t}w_t(\beta\cdot y_t\log(\hat{y}_t) + (1-y_t)\log(1-\hat{y}_t))  
$$
where $\beta$ represents \textbf{label-based weight} which typically has a value greater than 1, and $w_t = \dfrac{1}{Z_t} \exp(\frac{\sum_{j=1}^H|x_t-x_{t-j}|^2}{2\sigma^2_{t,H}})$ is the \textbf{value-based weight}, $H$ is the number of points before point $t$, $\sigma^2_{t,H}$ is the variance of a window size $H$ before point $t$, $Z_t$ is a normalization term of weights over all points.

The adjustment of weight is essential to make the network be able to learn the patterns from anomalies and converge faster given the unbalanced nature of the problem. 

\subsection{Data Augmentation}\label{subsec:data-aug}

Data augmentation~\cite{deep:learning:book}, which generates artificial data for training, is an effective way to improve performance in deep learning, especially when the amount of the training data is limited. Currently, very few work has been done on data augmentation for time series data~\cite{data:aug:tsc:2018,Um:data:aug}. Note that the labeled data in time series anomaly detection is generally very limited. In this subsection, we present several practical and effective data augmentation techniques specifically designed for time series after our robust decomposition, in both time domain and frequency domain.  

\subsubsection{Time Domain}

\begin{figure}[!htb]
    \centering
    \includegraphics[width=\linewidth]{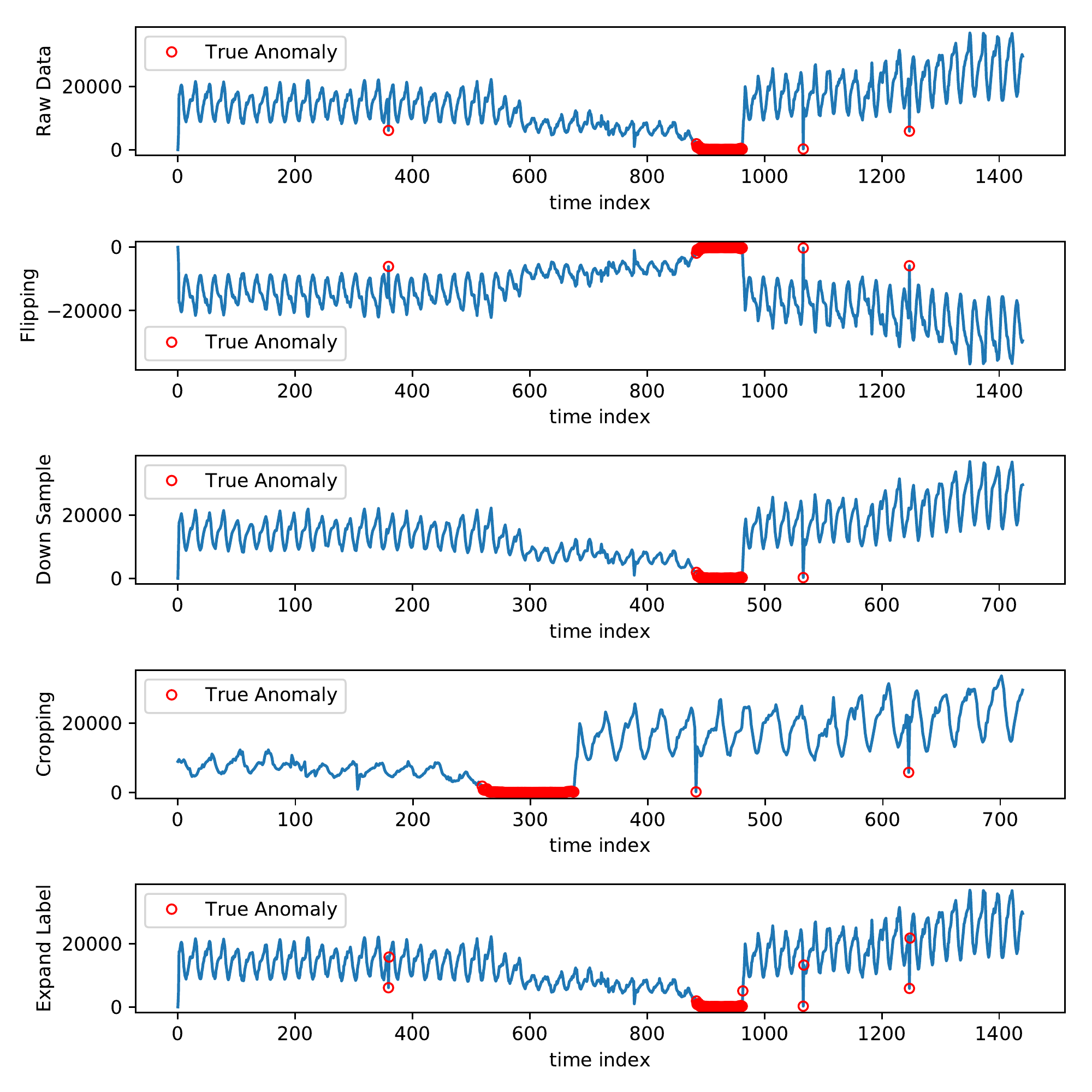}
    \caption{Plots of time series using time domain augmentations. The down sample rate is 2 and cropping length is half of the raw data length.}
    \label{fig:time_domain_augmetation}
\end{figure}

We summarize several effective transforms for time series anomaly detection in the time domain, including flipping, downsampling, cropping, and label expansion, etc. The augmented time series have been plotted in Figure~\ref{fig:time_domain_augmetation}. In the following discussion, we assume we have an input time series $x_1, \cdots, x_N$.  

\textit{Flipping.} We generate a new sequence $x^{'}_1, \cdots, x^{'}_N$ where $x^{'}_t = -x_t$ with the same anomaly labels. It can be used when we care anomalies in both directions. In the scenarios where we are interested in only one direction, this transform cannot be applied. 


\textit{Downsampling.} We down sample the original time series of length $N$ with a specific down sample rate $k$, to get a shorter time series with length $\lfloor N/k\rfloor$. 
The label series are also down sampled, or diluted, in the same rate $k$ as values series.

We have tried to duplicate and concatenate $k$ number of those shorter time series to form a new time series with the same length $N$ as the original time series. However, this method does not work well since the connecting points are usually big jumps and they break the time dependencies of time series.



\textit{Cropping.} We crop samples with replacement from the original time series of length $N$ to get shorter time series with length $N'$. The label series are also cropped, with the same time stamp as values series. This is similar to random crop in computer vision. 

\textit{Label Expansion.} In time series anomaly detection, the anomalies generally occur sequentially. As a result, a data point which is close to a labeled anomaly in terms of both time distance and value distance is very likely to be an anomaly. We select those data points and label as anomalies in our training dataset. 



\subsubsection{Frequency Domain}


To further increase the labeled data and utilize the special properties of time series data, we explore the data augmentation methods in the frequency domain. Specifically, we have developed several different policies, i.e., magnitude augmentation and phase augmentation, to generate more artificial labeled data. 


For the input time series $x_1, \cdots, x_N$, we can get its frequency spectrum $F(\omega_k)$ through discrete Fourier transform as follows: 
\begin{equation*}
F(\omega_k) \!=\! \frac{1}{{N}} \!\!\sum_{t=0}^{N-1} \! x_te^{-j\omega_k t}\!=\!\Re[F(\omega_k)] + j\Im[F(\omega_k)], ~ k\!=\!0,1,\!\cdots\!, N\!-\!1
\end{equation*}
where $\Re[F(\omega_k)]$ and $\Im[F(\omega_k)]$ are the real and imaginary parts of the spectrum respectively, and $\omega_k= \frac{2\pi k}{N}$ is the angular frequency. Since $F(\omega_k)$ is complex valued, a more informative representation can be obtained by its amplitude and phase spectra~\cite{BLACKLEDGET200675}:
\begin{equation*}
F(\omega_k) = A(\omega_k) \exp[j\theta(\omega_k)],
\end{equation*}
where $A(\omega_k)$ is the amplitude spectrum defined as
\begin{equation*}
A(\omega_k)=|F(\omega_k)|=\sqrt{\Re^2[F(\omega_k)] + \Im^2[F(\omega_k)]},
\end{equation*}
and $\theta(\omega_k)$ is the phase spectrum defined as 
\begin{equation*}
\theta(\omega_k)=\tan^{-1}\frac{\Im[F(\omega_k)]}{\Re[F(\omega_k)]}.
\end{equation*}
Note that the input time series is real signal, the length of amplitude and phase spectra is $N' = \Bigl\lceil \frac{N+1}{2} \Bigr\rceil$.

In the frequency domain, our intuitive idea is to make perturbations in magnitude and phase in selected segments in the frequency domain. We define the selected segment length $K$ by the ratio $r$:
$$
K=rN'.
$$
Let define the number of segments as $m_K$. Then, each segment with starting point as $k_i$ in frequency domain is obtained as:
$$
[k_i, k_i + K), ~\text{where}~ k_i \sim U(0, N'-K), ~ i=1,2,\cdots, m_K.
$$
Here we also ensure that the overlapping part of consecutive segments is not exceeding the half length of segment length $K$, i.e., 
$$
|k_i-k_{i+1}|\geq \frac{K}{2}, ~  i=1,2,\cdots, m_K-1.
$$

In magnitude augmentation, we make perturbations in the magnitude spectrum. Specifically, we replace the values of all points in the selected segment with $\upsilon$, where $\upsilon$ has Gaussian distribution as $\upsilon  \sim N(\mu_A,q_A \bar{\delta}^2_A)$,
where $\mu_A$ can be set as zero or $\bar{\mu}_A$ based on configuration, and $\bar{\mu_A}, \bar{\delta}^2_A$ is the sample mean and variance of the time series in the segment, respectively, and $q_A$ is adopted to control the degree of perturbation.

Similarly, we can make perturbation in the phase spectrum in phase augmentation. Specifically, the values of all points in the selected segment are increased by a small perturbation $\theta$, which is sampled from Gaussian distribution as
$\theta \sim N(0,\delta_{\theta}^2)$.

\begin{figure}[]
    \centering
    \includegraphics[width=1\linewidth]{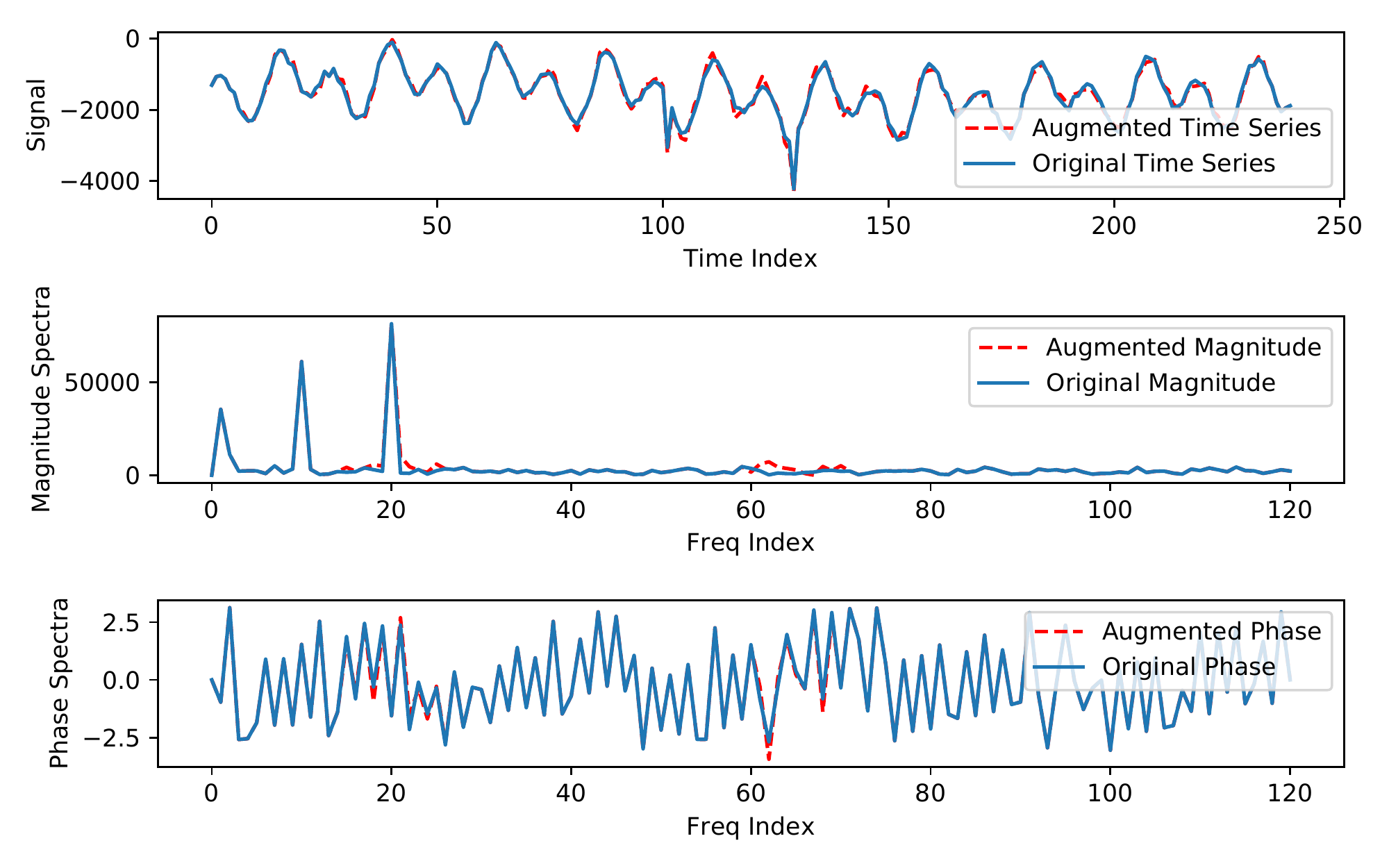}
    \caption{Plots of time series using magnitude augmentation and phase
    augmentation.}
    \label{fig:3-freqAugDemo}
\end{figure}

The proposed frequency-domain time series augmentation methods on a sample data is plotted in Figure~\ref{fig:3-freqAugDemo} for illustration.

\subsection{Online Inference for Streaming Time Series}

In the online inference stage, we need to perform both decomposition and the inference of the deep network.

For our decomposition algorithms RobustSTL and RobustTrend, we developed an online version to process the streaming data efficiently. Due to the space limit, we only present some details of the trend extraction, the most computationally expensive step here. Firstly, we assume that the change of trend is slow most of the time. Hence, we do not need to re-estimate trend every time a new point streams. We pick a small step size $q$ and only solve the optimization problem every $q$ data points. When trend has an abrupt change, there will be a delay at most $q$ points to estimate it in detecting such an anomaly, which is typically not a concern for users especially when $q$ is quite small (e.g., $q=5$). Secondly, when we need to solve the optimization problem to estimate trend difference $\nabla\tau$, we only consider data of length $\hat{w}$ instead of the whole time series of length $w$. Additionally, we use the ``warm start'' technique where the whole trend estimated from time $t-1$ is used as the initial solution when solving the optimization problem at time $t$, leading to a faster convergence speed. 

Although training the network with large amount of augmented samples is time-consuming, the online inference of the neural network is quite efficient as only matrix multiplications are needed. We are aware as time series stream and keep on coming, the new patterns might show up. Hence, we can update the trained model regularly to encode the latest information.

\section{System Architecture}

\begin{figure*}[]
    \centering
    \includegraphics[width=.8\linewidth]{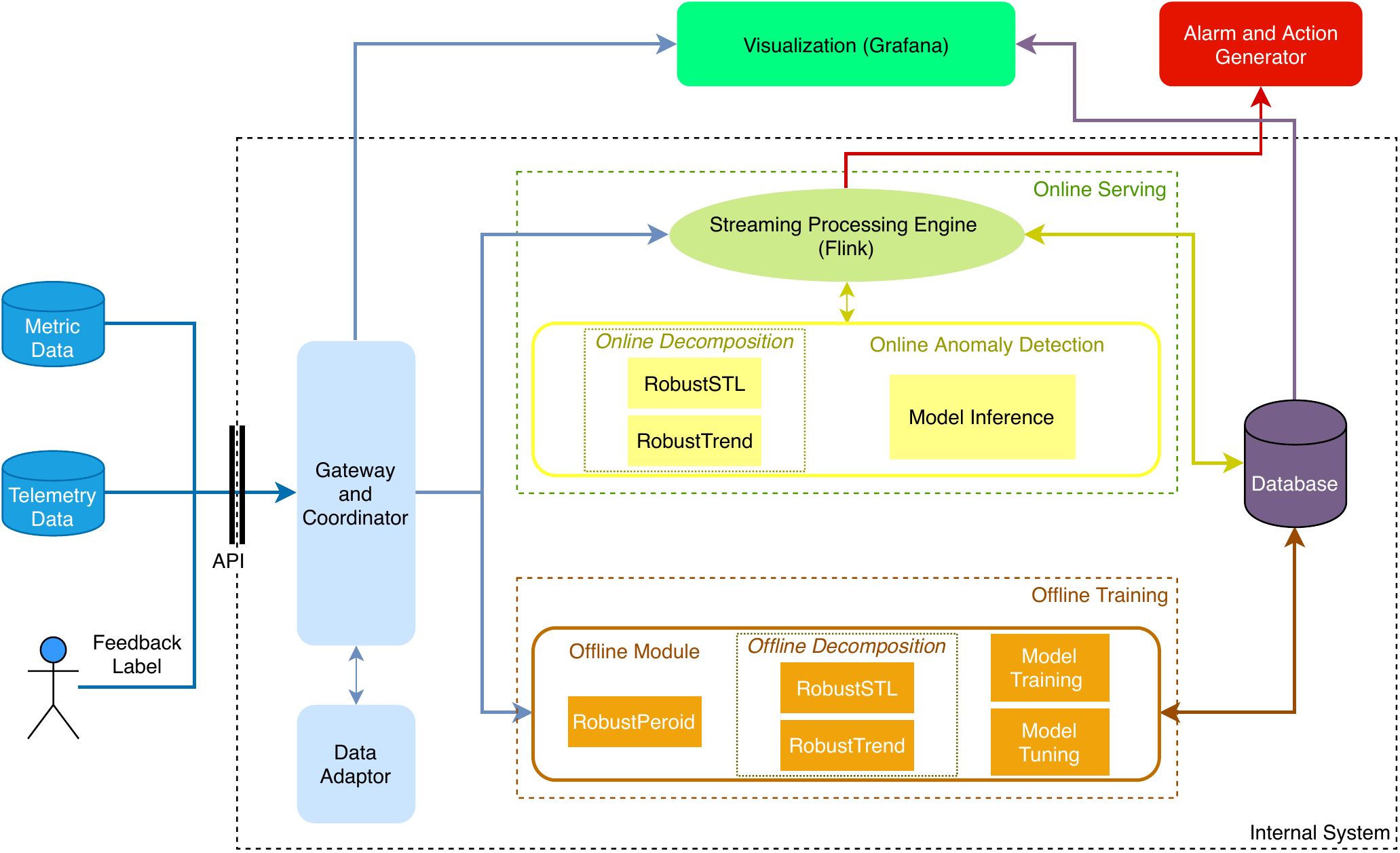}
    \caption{System architecture of RobustTAD framework.}
    \label{fig:3-sys-arch}
\end{figure*}

The system deployed at Alibaba Group consists of four major components: 1) data ingestion; 2) offline training; 3) online service; 4) visualization and alarms. Figure~\ref{fig:3-sys-arch} illustrates the general data flow pipeline of the system. Data ingestion allows users to provide both metric data and telemetry data from different sources, such as IoT devices, time series databases (TSDB), etc. In addition, user feedback labels can also be imported. The data is transmitted through an application program interface (API) to a coordinator, then a time series anomaly detection model can be trained in the offline training module. After training, the trained model will be sent to the database for future use. The heart of the system is a streaming processing engine based on Flink in the online service component. It reads the data in stream, and calls online RobustSTL/RobustTrend and online neural network inference modules to detect whether the current point is anomaly or not in real time. Note that both offline training and online service modules need to communicate with the database module regularly. When the latency becomes a main concern during online service, in memory database such as Redis and local caching mechanism could be used instead of the traditional relational database such as MySQL. Outside the internal anomaly detection system, there is also a visualization module that provides the dashboard for real-time monitoring of the time series, as well as an alarming module which notifies users the abnormal events once an anomaly is detected. 

During deploying the system to real world production environment, we observe that the offline training with large amount of augmented samples takes notable time on a GPU machine. For example, when we trained Yahoo data (described in subsection~\ref{subsubsec:yahoo}) on a V100 GPU card, it took about two hours. However, once the trained model is ready to be served for online inference, the computation cost is extremely low with less than 10 ms to predict the anomaly label for a new point. Including the online robust decomposition for time series adds some extra overheads with about 100 ms. Overall, the inference of a point being anomaly or not for streaming data could be done quite efficiently, which makes the approach suitable for many time series anomaly detection tasks in practice.



\section{Empirical Results}







\subsection{Datasets} \label{subsubsec:yahoo}
We use 367 time series from public Yahoo benchmark datasets~\cite{AD_laptev2015generic}. These time series are hourly sampled data with a duration of two months, which are collected from the real production traffic to some of the Yahoo website properties and better represent the complex patterns such as seasonality and trends in real-world time series data. The detailed properties are shown in Table~\ref{tab:Yahoo_description}. Yahoo data set has a good coverage of different varieties of anomalies in time series, such as seasonality, level change, variance change and their combinations.

\begin{table}[!h]
\centering
\caption{Characteristics of Yahoo benchmark datasets. Len: average length of the time series, Cnt:
number of series, TG: time granularity, AR: anomaly rate, SP: seasonal period, LS: level shift, VC: variance change. A1 are real operations series; A2-A4 are synthetic operations series. }

\label{tab:Yahoo_description}

\begin{tabular}{cccccccccc}
\hline

Dataset & Len & Cnt & TG  & AR & SP  & LS & VC \\ 

\hline
A1  & 1440   & 67  & 1 hr     & 1.76\% & 24,168  & Yes & Yes  \\
A2  & 1440   & 100 & 1 hr     & 0.33\% & 100-300 & Yes & Yes  \\
A3  & 1680   & 100 & 1 hr     & 0.56\% & 12,24,168 & Yes & Yes  \\
A4  & 1680   & 100 & 1 hr     & 0.62\% & 12,24,168  & Yes & Yes  \\ 
\hline

\end{tabular}
\end{table}

\subsection{Evaluation Metrics} \label{subsec:metric}
We compute the following evaluation metrics: 
\begin{itemize}
    \item Precision: $\frac{TP}{TP+FP}$, represents the ratio of TP in all the data points which are predicted/detected as anomaly.
    \item Recall: $\frac{TP}{TP+FN}$, represents the ratio of TP in the all the data points which are truly anomaly. 
    \item F1 score: $2\times\dfrac{Precision\cdot Recall}{Precision+Recall}$, an overall measure of accuracy that combines both precision and recall. Perfect accuracy is achieved when F1 is 1, and 0 otherwise.
\end{itemize}

In addition to the metrics mentioned above, we also adopt a relaxed version of F1 score~\cite{Choudhary2017-benchmark}, which could serve better to the test datasets where pattern anomalies exist. Pattern anomalies happen when a segment of consecutive points are labeled as anomalies. Instead of requiring an exactly point-wise match between a true anomaly and a detected anomaly for a point to be counted towards TP, it is reasonable to give some leeway by allowing a lag up to window size $m$. This corresponds to the scenario where the detected anomaly points might be earlier or delayed by a few points compared with the labeled anomaly points. We would still count these detected points as TP. We choose $m=3$ to allow a mismatch up to 3 adjacent points. During the experiment, TP, FP, FN will be aggregated over multiple time series for one dataset and be used to produce a single micro-precision/recall/F1 score. This gives more weights to the time series labeled with more anomalies and better presents the average performance of each method.


\subsection{Experimental Setup}
The following state-of-the-art methods are used for comparison:
\begin{itemize}
    \item \textit{ARIMA}\footnote{https://www.rdocumentation.org/packages/forecast/versions/8.5/topics/auto.arima}: ARIMA models~\cite{TSanalysisBook_box2015time} are well known for its versatility to forecast a time series. We label a prediction point as anomaly if its true value lies outside the prediction intervals;
    \item \textit{SHESD}\footnote{https://github.com/twitter/AnomalyDetection}: SHESD~\cite{twitter:anomaly:detection} is an extension of both ESD and S-ESD. It decomposes and  extracts out the seasonal component, then applies robust statistics median and MAD to calculate anomaly score;
    \item \textit{Donut}\footnote{https://github.com/haowen-xu/donut}:  Donut~\cite{xu2018unsupervised} is an unsupervised, variational auto-encoder (VAE) based algorithm for detecting anomalies in seasonal KPIs in DevOps domain. It has competitive experimental results and a solid theoretical explanation.
\end{itemize}

In additional to these methods, we also compare the performance of different U-Net settings:
\begin{itemize}
    \item \textit{U-Net-Raw}. Apply U-Net to raw time series data directly.
    \item \textit{U-Net-De}. Apply U-Net to decomposed remainder.
    \item \textit{U-Net-DeW}. Adopt weight adjusted loss for U-Net-De.
    \item \textit{U-Net-DeWA}. Adopt data augmentation for U-Net-DeW.
\end{itemize}



During our experiments, we split each time series into left half and right half, corresponding to the train part and test part. We feed the training parts from all samples to train the network/model together. Then we predict and test the results using all the test parts. During the training and testing, we use online mode, e.g. the sliding window mode, with window size at 240 points. In each sliding window, the model predicts the label of last point or right most point, then progresses one point at a time to the right. To have a relative fair comparison, for unsupervised algorithms, labels in the training part are used to tune hyper parameters, such as probability threshold, which is chosen by matching predicting scores with the true labels to maximize F1 within training data.

\subsection{Results}



Table~\ref{tab:5-alg-compare} shows the performance comparison of different methods, including both F1 score and the relaxed version. SHESD achieves acceptable result. Donut, as it was designed for DevOps data, has good recall but low precision. For the U-Net-based architecture, we can see that a naive version of the U-Net without any adjustments only lead to a 0.403 F1 score and 0.533 relaxed F1 score. However, once we include the decomposition, we can see a significant 0.22 increase in F1. By adopting the aforementioned loss adjustment, as well as augmentation, we are able to further increase F1 up to 0.693, and relaxed F1 up to 0.812, which is far better than the previous STOA. Given the relatively high F1 score of SHESD and U-Net-DeWA, we also take a closer look at the performance of these two methods on each Yahoo subset in Figure~\ref{tab:5-yahoo-subset-perf}, which further demonstrated the supreme performance of our proposed approach on both real world operations series, as well as synthetic series.

\begin{table}[!h]
    \centering
        \caption{Performance comparison of different methods. Acronyms in U-Net section: De - decomposition, W - weight adjustment, A - augmentation}
\begin{tabular}{ccccc}
\hline
{} &  Precision &  Recall &  F1 Score &  Relax F1 Score \\
\hline
ARIMA      &      0.513 &   0.144 &     0.225 &           0.533 \\
SHESD      &      0.501 &   0.488 &     0.494 &           0.557 \\
Donut      &      0.015 &   0.829 &     0.029 &           0.030 \\
\hline
U-Net-Raw  &      0.473 &   0.351 &     0.403 &           0.533 \\
U-Net-De   &      0.651 &   0.594 &     0.621 &           0.710 \\
U-Net-DeW  &      0.793 &   0.569 &     0.662 &           0.795 \\
U-Net-DeWA &      0.859 &   0.581 &     \textbf{0.693} &           \textbf{0.812} \\
\hline
\end{tabular}
    \label{tab:5-alg-compare}
\end{table}

\begin{table*}[!h]
    \centering
        \caption{Performance of SHESD and U-Net-DeWA on each Yahoo subset.}
\begin{tabular}{c|cccc|cccc}
\hline
{} & \multicolumn{4}{c}{SHESD} & \multicolumn{4}{|c}{U-Net-DeWA } \\
{Dataset} & Precision & Recall & F1 Score & Relax F1 Score &  Precision & Recall & F1 Score & Relax F1 Score \\
\hline
A1 &     0.364 &  0.406 &    0.384 &          0.472 &      0.657 &  0.279 &    0.392 &          \textbf{0.625} \\
A2 &     0.548 &  0.285 &    0.375 &          0.643 &      0.955 &  0.950 &    0.952 &          \textbf{0.963} \\
A3 &     0.980 &  0.732 &    0.838 &          0.838 &      0.996 &  1.000 &    0.998 &          \textbf{0.998} \\
A4 &     0.544 &  0.535 &    0.539 &          0.543 &      0.929 &  0.880 &    0.904 &          \textbf{0.904} \\
\hline
\end{tabular}
    \label{tab:5-yahoo-subset-perf}
\end{table*}

\begin{figure*}[!h]
 \centering
\begin{subfigure}[b]{0.4\textwidth}
\includegraphics[width=1\textwidth]{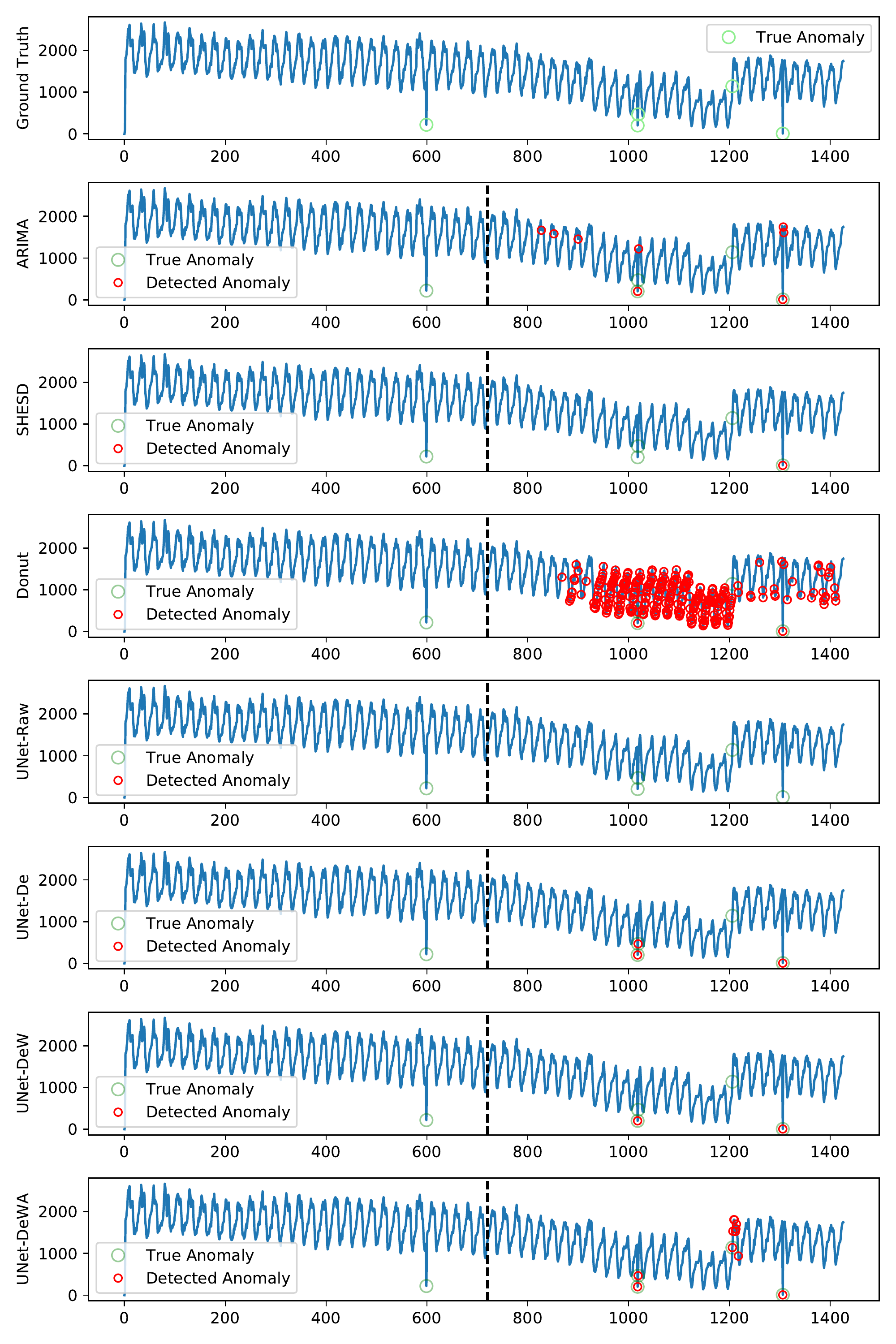} 
\caption{Plot of raw values with detected anomaly.}
\label{fig:5-yahoo-real}
\end{subfigure}
\begin{subfigure}[b]{0.4\textwidth}
\includegraphics[width=1\textwidth]{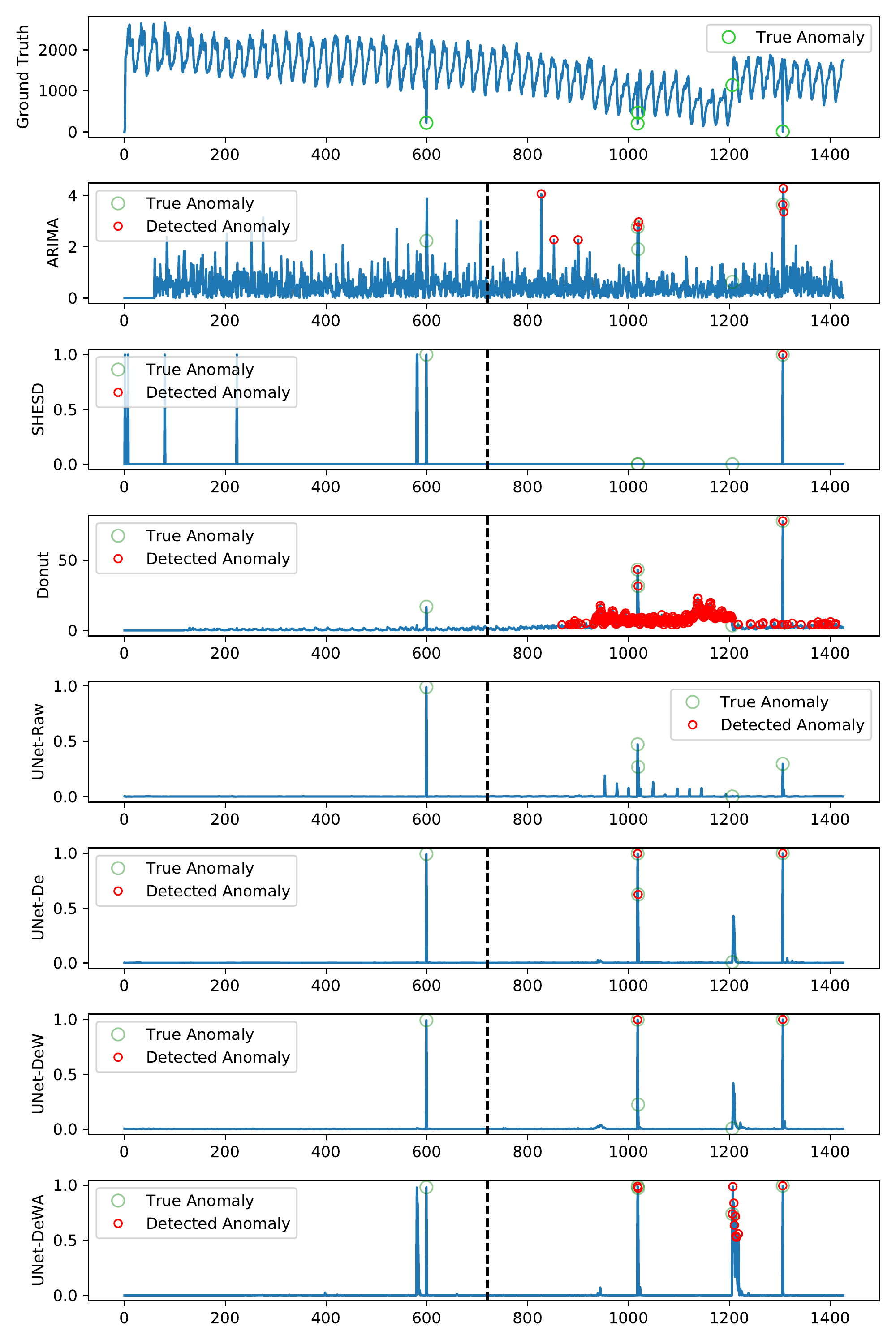}
\caption{Plot of anomaly scores with detected anomaly.}
\label{fig:5-yahoo-real-score}
\end{subfigure}
\caption{Performance of each method on a specific time series. The values before the vertical black dashed line represent the training data, and those after represent the testing data.}
\label{fig:pef}
\end{figure*}

To better understand how each method performs on a specific time series, we demonstrate one real world data from Yahoo A1 in Figure~\ref{fig:pef}, where it includes seasonality combined with level changes and spikes. The original U-Net-Raw fails to detect anomalies in this time series. However, as we include decomposition, loss adjustment, and data augmentation in our framework, the network has demonstrated learning capabilities, being able to identify those anomalies correctly in Figure~\ref{fig:5-yahoo-real}. We further examined the anomaly scores generated by each method in Figure~\ref{fig:5-yahoo-real-score}. As we can see, for ARIMA and Donut, an appropriate threshold needs to be picked in order to identify anomalies correctly. On the other hand, the network basically outputs the probability that each point being anomaly or not. A fixed threshold 0.5 can be used to detect anomaly. For the U-Net-Raw, the probability score is below 0.5 during testing, leading to many false negatives. Meanwhile, as we apply the adjustments we discussed in Section~\ref{subsec:framework}, more anomalies are being identified correctly. It is worth mentioning that for the anomalies showing at index position 1200 where there is a level change combined with seasonality, U-Net-De and U-Net-DeW all fail to identify correctly despite the probabilities of being anomaly is very close to 0.5. On the other hand, as we incorporate data augmentation from both time domain and frequency domain, such anomalies could be easily identified with U-Net-DeWA.


\section{Discussion and Conclusion}

The empirical comparison on the Yahoo datasets clearly demonstrates the promising performance of encoder-decoder structure for time series anomaly detection. In particular, it can effectively extract multi-scale features. Intuitively, in anomaly detection we need to compare each observation with its context to determine its label. The multi-scaled features learned actually describe the context from different levels quite well, thus leading to good anomaly detection performance. Also note that the general framework integrating both time series decomposition and deep network with some modifications can be applicable for many other time series tasks such as forecasting.  



In summary, in this paper we propose RobustTAD, a Robust Time series Anomaly Detection framework based on time series decomposition and convolutional neural network for time series data. Combined with both of their advantages, this algorithm dose not only achieve high performance and high efficiency, but also possess the ability to to handle complicated patterns and lack of sufficient labels, which makes it practical and effective approach to serve cloud and IoT monitoring. 


More future work can be done to extend RoubstTAD, including: 1) explore more network architectures for better learning of multi-scale features. For example, we can impose a loss function on each decoder layer~\cite{Hou2017} for a better representation learning; 2) multi-channel study by feeding three decomposed components directly to the network (in other words, $C=3$ in Figure~\ref{fig:3-net-arch}); 3) a joint-learning framework to learn the time series decomposition and anomaly labels at the same time. Similar work has been done on image segmentation, such as Faster R-CNN~\cite{Ren2017} and Mask R-CNN~\cite{he2017mask}. It is worth exploring the possibilities of jointly learning the characteristics of time series as well.



%
\bibliographystyle{ACM-Reference-Format}
\bibliography{bibliography}

%

\end{document}